# A Differential Semantics of Lazy AR Propagation

**Anders L Madsen**
HUGIN Expert A/S
Gasværksvej 5
9000 Aalborg, Denmark
Anders.L.Madsen@hugin.com

## Abstract

In this paper we present a differential semantics of Lazy AR Propagation (LARP) in discrete Bayesian networks. We describe how both single and multi dimensional partial derivatives of the evidence may easily be calculated from a junction tree in LARP equilibrium. We show that the simplicity of the calculations stems from the nature of LARP. Based on the differential semantics we describe how variable propagation in the LARP architecture may give access to additional partial derivatives. The cautious LARP (cLARP) scheme is derived to produce a flexible cLARP equilibrium that offers additional opportunities for calculating single and multi dimensional partial derivatives of the evidence and subsets of the evidence from a single propagation. The results of an empirical evaluation illustrates how the access to a largely increased number of partial derivatives comes at a low computational cost.

## 1 INTRODUCTION

Bayesian networks (BNs) (Pearl 1988; Cowell, Dawid, Lauritzen, and Spiegelhalter 1999; Jensen 2001) are efficient knowledge representations for reasoning under uncertainty. Usually, the inference process in a BN proceeds either by message passing in a secondary computational structure (Lauritzen and Spiegelhalter 1988; Shenoy and Shafer 1990; Jensen, Lauritzen, and Olesen 1990) or by direct computations (Shachter 1986; Pearl 1988; Li and D'Ambrosio 1994; Zhang and Poole 1994; Dechter 1996; Darwiche 2003).

Darwiche (2003) gives a differential semantics of inference by considering it as the task of computing the value of an algebraic function while Park and Darwiche (2004) based on a similar algebraic function representation give a differential semantics of the HUGIN (Jensen, Lauritzen, and Olesen 1990) and Shenoy-Shafer (Shenoy and Shafer 1990) algorithms.

In this paper we present a differential semantics of Lazy Propagation (LP) (Madsen and Jensen 1999). LP is another join-tree-based algorithm for inference in BNs. The main motivation for this work is to determine the flexibility and suitability of LP w.r.t. computing partial derivatives of the evidence. Computing partial derivatives is an important part of hypothesis driven parameter sensitivity analysis, for instance.

We show how the fundamental properties of LP — maintain decompositions of clique and separator potentials, postpone the elimination of variables for as long as possible, take advantage of barren variables, and exploit independence relations induced by evidence — support efficient calculation of single and multi dimensional partial derivatives of the evidence and subsets of the evidence from a junction tree in equilibrium. Due to space limitations we focus on LARP (Madsen 2004b) where messages are computed using arc-reversal (Shachter 1986). In addition we describe how variable propagation (Jensen 2001) and a cautious propagation scheme (Jensen 1995) support the calculation of additional derivatives. The paper contains an empirical evaluation of the overhead introduced by the cautious propagation scheme.

## 2 PRELIMINARIES AND NOTATION

A discrete BN $\mathcal{N} = (\mathcal{X}, \mathcal{G}, \mathcal{P})$ over variables $\mathcal{X}$ consists of an acyclic, directed graph $\mathcal{G} = (V, E)$ and a set of conditional probability distributions (CPDs) $\mathcal{P} = \{\Theta_{X \mid \pi(X)} : X \in \mathcal{X}\}$. The vertices $V$ of $\mathcal{G}$ correspond one-to-one with the variables of $\mathcal{X}$, i.e. $\mathcal{X} \sim V$. $\mathcal{N}$ induces a factorization of the joint probability distribution over $\mathcal{X}$:

$$P(\mathcal{X}) = \prod_{X \in \mathcal{X}} \Theta_{X \mid \pi(X)}, \qquad (1)$$

where $\pi(X)$ is the set of variables corresponding to the parents of the vertex representing $X$ in $\mathcal{G}$, $fa(X) = \pi(X) \cup \{X\}$, and $\Theta_{X|\pi(X)}$ is the CPD of $X$ given its parents $\pi(X)$. We use $\theta_{x|\pi}$ to denote the entry of $\Theta_{X|\pi(X)}$ where $X = x$ and $\pi(X) = \pi$, i.e. $(x, \pi)$ is a configuration of $(X, \pi(X))$ and $\theta_{x|\pi} = \Theta_{X=x|\pi(X)=\pi}$. We denote a configuration of $(X_i, \pi(X_i))$ as $(x_i, \pi_i)$

A probability potential $\phi(X|Y)$ where $X, Y \subseteq \mathcal{X}$ is a non-negative function over $X \cup Y$. Let $\phi(H|T)$ be a probability potential with head $H = H(\phi)$ and tail $T = T(\phi)$, then the domain $dom(\phi)$ of $\phi$ is defined as $dom(\phi) = H \cup T$. When the distinction between head and tail variables is not important, we will indicate the domain $W = H \cup T$ of $\phi$ as $\phi_W$. The domain $dom(\Phi)$ of a set of potentials $\Phi = \{\phi_1, \ldots, \phi_n\}$ is defined as $dom(\Phi) = \bigcup_{i=1}^{n} dom(\phi_i)$.

For evidence $\epsilon_X = \{X = x\}$ on $X$, an evidence function $f(X)$ takes on values $f(x) = 1$ and $f(y) = 0$ for all $y \neq x$ and we use $f(x)$ as an abbreviation for $f(X = x)$. If $X$ is not observed, then $f(x) = 1/|X|$. Hard evidence $\epsilon_X = \{X = x\}$ enables us to instantiate all potentials $\phi$ with $X \in dom(\phi)$ to reflect $X = x$ reducing the domain size of $\phi$ by one. We refer to the instantiation of $\phi$ as an application of $f(X)$ on $\phi$. The instantiation of potentials enables us to take advantage of independence relations induced by $\epsilon$.

## 3 POTENTIALS AND OPERATIONS

Probabilistic inference using LARP proceeds by message passing in a junction tree $\mathcal{T} = (\mathcal{C}, \mathcal{S})$ where sets of probability potentials are passed between cliques $\mathcal{C}$ and separators $\mathcal{S}$. Messages are passed in two flows (collect and distribute) relative to a preselected root $R \in \mathcal{C}$. We define the notion of a clique / separator potential (referred to as a potential).

**Definition 3.1 [Potential]**
A *potential* on $W \subseteq \mathcal{X}$ is a singleton $\pi_W = (\Phi)$ where $\Phi$ is a set of non-negative real functions on subsets of $W$.

We call a potential $\pi_W$ *vacuous* if $\pi_W = (\emptyset)$. The vacuous potential is denoted $\pi_\emptyset$. We define the operations of combination, division, and contraction as follows:

**Definition 3.2 [Combination]**
The *combination* of potentials $\pi_{W_1} = (\Phi_1)$ and $\pi_{W_2} = (\Phi_2)$ denotes the potential on $W_1 \cup W_2$ given by:

$$\pi_{W_1} \otimes \pi_{W_2} = (\Phi_1 \cup \Phi_2).$$

**Definition 3.3 [Division]**
The *division* of potentials $\pi_{W_1} = (\Phi_1)$ and $\pi_{W_2} = (\Phi_2)$ denotes the potential on $W_1$ given by:

$$\pi_{W_1} \ominus \pi_{W_2} = (\Phi_1 \setminus \Phi_2).$$

Notice the simplicity of potential combination and division. The operations reduce to set manipulations.

**Definition 3.4 [Contraction]**
The contraction $c(\pi_W)$ of a potential $\pi_W = (\Phi)$ is the non-negative function on $W$ given by:

$$c(\pi_W) = \prod_{\phi \in \Phi} \phi.$$

We define the contraction of $\pi_\emptyset$ as $c(\pi_\emptyset) = 1$.

## 4 LAZY AR PROPAGATION

Inference in LARP proceeds, as mentioned, by message passing in a junction tree representation $\mathcal{T} = (\mathcal{C}, \mathcal{S})$ of $\mathcal{N}$. Prior to the message passing $\mathcal{T}$ is initialized. During message passing variables are eliminated by marginalization. Marginalization proceeds by arc-reversal and barren variable elimination.

### 4.1 INITIALIZATION

The first step in initialization of $\mathcal{T}$ is to associate a vacuous potential with each clique $A \in \mathcal{C}$. Initialization proceeds by assigning $\pi_X = (\{\Theta_{X|\pi(X)}\})$ for each $X \in \mathcal{X}$ with a clique $A$, which can accommodate it, i.e. $fa(X) \subseteq A$.

After initialization each clique $A$ holds a potential $\pi_A = (\Phi)$. The set of clique potentials is invariant during insertion and propagation of evidence. After initialization the joint potential $\pi_\mathcal{X}$ on $\mathcal{T} = (\mathcal{C}, \mathcal{S})$ is:

$$\pi_\mathcal{X} = \bigotimes_{A \in \mathcal{C}} \pi_A = \left( \bigcup_{X \in \mathcal{X}} \{\Theta_{X|\pi(X)}\} \right).$$

Let $\pi_A = (\Phi)$ be the clique potential for clique $A$. The domain of the contraction of $\pi_A$ is:

$$dom(c(\pi_A)) = \bigcup_{\phi \in \Phi} dom(\phi)$$

and has the property $dom(c(\pi_A)) \subseteq A$.

Subsequently, an evidence function $f(X)$ is assigned to each $A$ with $X \in A$ for all $X$. If $\{f_1, \ldots, f_{|A|}\}$ is the set of evidence functions associated with $A$, then $\pi_{\epsilon_A} = (\{f_1, \ldots, f_{|A|}\})$ is the evidence potential for clique $A$.

### 4.2 MESSAGES

The message $\pi_{A \to B}$ is passed from clique $A$ to clique $B$ by absorption, see Figure 1. Absorption from $A$ to $B$ involves eliminating the variables $A \setminus B$ from the combination of the potential associated with $A$ and the

messages passed to A from adjacent cliques adj(A) except B. In principle $\pi_{A \to B}$ is computed as:

$$\pi_{A \to B} = \left(\pi_A \otimes \pi_{\epsilon_A} \otimes \left(\otimes_{C \in \text{adj}(A) \setminus \{B\}} \pi_{C \to A}\right)\right)^{\downarrow B}, \tag{2}$$

where $\pi_{C \to A}$ is the message passed from C to A. In LARP the projection operation of (2) proceeds as:

1. Let $\Phi$ be the potentials of $\pi$ where:

$$\pi = \pi_A \otimes \pi_{\epsilon_A} \otimes \left(\otimes_{C \in \text{adj}(A) \setminus \{B\}} \pi_{C \to A}\right).$$

2. Remove from $\Phi$ potentials of barren variables.

3. If $X \in A$ is observed, then apply $f(X)$ on potentials of $\Phi$.

4. Remove from $\Phi$ each potential $\phi$ where $\text{dom}(\phi)$ is separated from B by $\epsilon$.

5. Eliminate each variable $X \notin S$ from $\Phi$ by a sequence of arc-reversals and barren variable eliminations. Let $\Phi^*$ be the result.

6. Associate the message $\pi_{A \to B} = (\Phi^*)$ with S.

Each message $\pi_{A \to B}$ has the form $\pi_{A \to B} = (\{\phi_1, \ldots, \phi_n\})$ where $|H(\phi_i) = 1|$ for all i. A head variable may be instantiated by evidence though. Details may be found in (Madsen 2004a; Madsen 2004b).

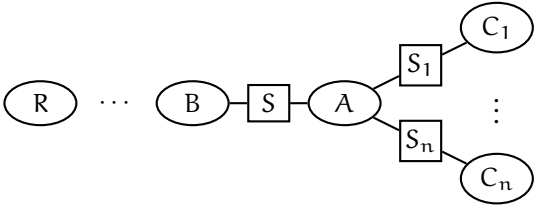

Figure 1: A junction tree with root clique R.

Notice that in the traditional LP scheme all evidence may not be received by each clique. Consider Figure 1 and assume B is instantiated by evidence. In this case, any evidence located in cliques to the left of B will not be included in the message passed from B to A. We make sure to propagate all evidence by skipping step 4. This may, however, decrease the performance of evidence propagation as additional calculations are required. In Section 8 we report on an empirical evaluation of the impact of this adjustment.

### 4.3 POSTERIOR CLIQUE POTENTIAL

After completion of the two-phase message passing the potential $\pi_A^*$ of each clique A is:

$$\pi_A^* = (\Phi_A^*) = \pi_A \otimes \pi_{\epsilon_A} \otimes \left(\otimes_{C \in \text{adj}(A)} \pi_{C \to A}\right),$$

where $\pi_{C \to A}$ is the message received from $C \in \text{adj}(A)$, $\pi_A$ is the potential associated with A, and $\pi_{\epsilon_A}$ is the evidence potential associated with A.

The joint probability distribution $P(A, \epsilon)$ is obtained as the contraction $c(\pi_A^*)$ of $\pi_A^*$:

$$P(A, \epsilon) = c(\pi_A^*) = \prod_{\phi \in \Phi_A^*} \phi.$$

Notice that $\pi_X$ remains invariant during message passing. Finally, we have:

$$P(\epsilon) = \sum_A P(A, \epsilon) = \sum_A c(\pi_A^*) = \sum_A \prod_{\phi \in \Phi_A} \phi.$$

## 5 A DIFFERENTIAL SEMANTICS

We present a differential semantics of LARP similar to the semantics of HUGIN and Shenoy-Shafer propagation given by Park and Darwiche (2004). Park and Darwiche (2004) give the following differential semantics of $P(x|\epsilon)$, $P(\epsilon \setminus \epsilon_X)$ and $P(x, \pi|\epsilon)$:

$$P(x|\epsilon) = \frac{1}{P(\epsilon)} \frac{\partial P(\epsilon)}{\partial f(x)},$$

where $X \notin \epsilon$ and $f(X)$ is the (uniform) evidence function for X,

$$P(\epsilon \setminus \epsilon_X) = \sum_x \frac{\partial P(\epsilon)}{\partial f(x)},$$

where $X \in \epsilon$, and

$$P(x, \pi|\epsilon) = \left[\frac{\theta_{x|\pi}}{P(\epsilon)} \frac{\partial P(\epsilon)}{\partial \theta_{x|\pi}}\right](x, \pi),$$

where $\pi$ is a configuration of $\pi(X)$.

Park and Darwiche (2004) show how the above probabilities and the partial derivatives may be computed from either a HUGIN or a Shenoy-Shafer propagation.

### 5.1 SINGLE DIMENSIONAL PARTIAL DERIVATIVE

Let $\mathcal{T} = (\mathcal{C}, \mathcal{S})$ be a consistent junction tree with $A \in \mathcal{C}$ and let $\pi_A^* = (\Phi_A^*)$ where $\Phi_A^* = \{\phi_1, \ldots, \phi_n\}$. Each potential $\phi \in \Phi_A^*$ is either a CPD $\Theta_{X|\pi(X)} \in \mathcal{P}$, an evidence function $f(X)$, or a probability potential $\phi = \sum_W \prod_{\phi' \in \Phi'} \phi'$ where $\Phi' \subseteq \mathcal{P}$ and $W \subseteq \left(\bigcup_{\phi' \in \Phi'} \text{dom}(\phi')\right) \setminus \text{dom}(\phi)$. CPDs and evidence functions are associated with A as part of the initialization of $\mathcal{T}$ whereas CPDs and probability potentials are received by A from adjacent cliques during inference. For each $X \in A$ s.t. $\text{fa}(X) \subseteq A$ we may compute:

$$\frac{\partial P(\epsilon)}{\partial \theta_{x|\pi}} = \left[\sum_{A \setminus \text{fa}(X)} c(\pi_A^* \ominus \pi_X)\right](x, \pi), \tag{3}$$

where $\pi_X = (\{\Theta_{X\mid \pi(X)}\})$ and $\pi$ is a configuration of $\pi(X)$. Notice that $A$ may be any clique for which $fa(X) \subseteq A$. Similarly, we may compute:

$$\frac{\partial P(\epsilon)}{\partial f(x)} = \left[\sum_{A\setminus\{X\}} c(\pi_A^* \ominus \pi_{f(X)})\right](x). \quad (4)$$

where $\pi_{f(X)} = (\{f(X)\})$. Notice that $A$ may be any clique for which $X \in A$.

The partial derivative of $P(\epsilon)$ w.r.t. a parameter $\theta_s$ of the contraction of a message $\pi_{C_i \to A}$ is:

$$\frac{\partial P(\epsilon)}{\partial \theta_s} = \left[\sum_{A\setminus S} c(\pi_A^* \ominus \pi_{C_i \to A})\right](s), \quad (5)$$

where $C_i \in adj(A)$ is a clique adjacent to $A$, $S = A \cap C_i$, and $s$ is a configuration of $S$.

Notice that decompositions of the derivatives in (3) - (5) are obtained if potential marginalization is applied instead of contraction.

The key difference between the Shenoy-Shafer and LP is that in the latter messages are factorized into sets of potentials. This factorization offers some additional opportunities with respect to computing partial derivatives. That is, the decomposition of clique and separator potentials can be exploited to increase the number of single dimensional partial derivatives, which may be computed after a single propagation. We are able to compute additional single dimensional derivatives w.r.t. evidence and single parameters.

In addition to computing $\frac{\partial P(\epsilon)}{\partial \theta_s}$ for any separator $S = A \cap C$ for $C \in adj(A)$, we may compute the partial derivative of $P(\epsilon)$ with respect to each potential $\phi$ received from a adjacent clique $C$ of $A$, i.e. $\phi \in \Phi$ for some message $\pi_{C \to A} = (\Phi)$. This is an extension compared to the differential semantics of Shenoy-Shafer and HUGIN propagation.

**Example 5.1**
*Due to the factorization of clique potentials and the use of arc-reversal for message computation, the partial derivative $\frac{\partial P(\epsilon)}{\partial \phi(x\mid u)}$ for any $\phi(X\mid U) \in \Phi_A^*$ where $\pi_A^* = (\Phi_A^*)$ is easy to compute:*

$$\frac{\partial P(\epsilon)}{\partial \phi(x\mid u)} = \left[\sum_{A\setminus(\{X\}\cup U)} c(\pi_A^* \ominus \pi_\phi)\right](x,u),$$

*where $\pi_\phi = (\{\phi(X\mid U)\})$. Notice $\phi$ is either a potential of $\mathcal{P}$ or a computed potential. In the later case, we have computed $\phi$ from a subset $\mathcal{P}_\phi$ of $\mathcal{P}$ by a sequence of arc-reversals and barren variable eliminations.*

## 5.2 MULTI DIMENSIONAL PARTIAL DERIVATIVE

Due to the decomposition of $\pi_A^*$ into the potentials $\Phi_A^*$, we may easily compute multi dimensional partial derivatives of $P(\epsilon)$ w.r.t. parameters, evidence functions, and potentials of received messages represented in $\Phi_A^*$ as well as a mixture of the three. For instance, we may from $\pi_A^*$ compute multi dimensional partial derivatives such as:

$$\frac{\partial^n P(\epsilon)}{\partial \theta_{x_1\mid \pi_1} \cdots \partial \theta_{x_k\mid \pi_k} \partial f(y_1)\cdots \partial f(y_l)\partial \phi(z_1)\cdots \partial \phi(z_m)}, \quad (6)$$

where $n = k + l + m$ and

$$\begin{aligned}
\Phi_\Theta &= \{\Theta_{X_1\mid \pi(X_1)},\ldots,\Theta_{X_k\mid \pi(X_k)}\} \subseteq \Phi_A^*, \\
\Phi_f &= \{f(Y_1),\ldots,f(Y_l)\} \subseteq \Phi_A^*, \\
\Phi_\phi &= \{\phi(Z_1),\ldots,\phi(Z_m)\} \subseteq \Phi_A^*.
\end{aligned}$$

Letting $\pi_\Theta = (\Phi_\Theta)$, $\pi_f = (\Phi_f)$, $\pi_\phi = (\Phi_\phi)$ we may compute (6) as:

$$\left[\sum_{A\setminus W} c(\pi_A^* \ominus \pi_\Theta \ominus \pi_f \ominus \pi_\phi)\right](w),$$

where $w$ is an instantiation of $W = dom(c(\pi_\Theta)) \cup dom(c(\pi_f)) \cup dom(c(\pi_\theta))$.

As a special case we may compute the multi dimensional partial derivative $\frac{\partial^n P(\epsilon)}{\partial \phi_1 \cdots \partial \phi_n}$ at clique $A$ where $\Phi_A^* = \{\phi_1,\ldots,\phi_n\}$ as:

$$\begin{aligned}
\left[\frac{\partial^n P(\epsilon)}{\partial \phi_1 \cdots \partial \phi_n}\right](a) &= \left[\sum_{A\setminus A} c(\pi_A^* \ominus \pi_{\Phi_A^*})\right](a) \\
&= [c(\pi_\emptyset)](a) = 1_a,
\end{aligned}$$

where $\pi_{\Phi_A^*} = (\Phi_A^*) = \pi_A^*$ and $1_a$ is function which returns 1 for all configuration $a$ of $A$.

**Example 5.2**
*Let $\mathcal{T} = (\mathcal{C}, \mathcal{S})$ be a junction tree in LARP equilibrium. Figure 2 shows the content of the messages passed to clique $A \in \mathcal{C}$ from adjacent cliques and the potential associated with $A$.*

*Assuming $\phi_1 = \Theta_{X_1\mid \pi(X_1)}, \phi_9 = \Theta_{X_9\mid \pi(X_9)} \in \mathcal{P}$ with $dom(\phi_1), dom(\phi_9) \subseteq A$ we may compute:*

$$\begin{aligned}
&\frac{\partial^2 P(\epsilon)}{\partial \theta_{x_1\mid \pi_1} \partial \theta_{x_9\mid \pi_9}} \\
&= \left[\sum_{A\setminus (fa(X_1)\cup fa(X_9))} c(\pi_A^* \ominus \pi_{X_1} \ominus \pi_{X_9})\right](x_1,\pi_1,x_9,\pi_9) \\
&= \left[\sum_{A\setminus (fa(X_1)\cup fa(X_9))} (\phi_{10}\phi_{11}\prod_{i=2}^{8} \phi_i)\right](x_1,\pi_1,x_9,\pi_9).
\end{aligned}$$

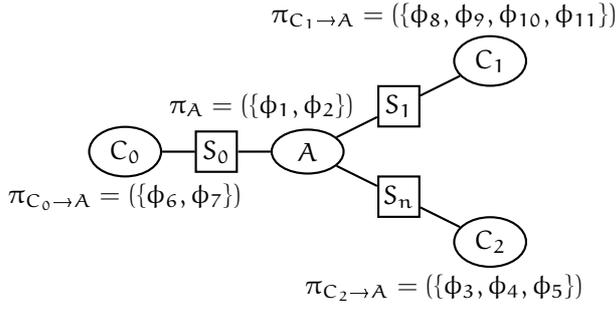

Figure 2: The content of the messages passed to A.

## 6 VARIABLE PROPAGATION

Variable propagation (Jensen 2001; Madsen and Jensen 1999) is a method for computing arbitrary joint probability distributions in a (consistent) junction tree $\mathcal{T} = (\mathcal{C}, \mathcal{S})$. The aim of variable propagation is to compute the joint probability distribution $P(W)$ for a set $W \not\subseteq C$ for any $C \in \mathcal{C}$. If $W \subseteq C$ for any $C \in \mathcal{C}$, then $P(W) = \sum_{C \setminus W} c(\pi_C^*)$.

Variable propagation using any variant of LP is particularly simple as it corresponds to *not* performing certain computations as computations are only performed when variables are eliminated.

Let $\mathcal{T} = (\mathcal{C}, \mathcal{S})$ be a consistent junction tree and consider the calculation of the joint potential $\pi_{A \cup \{X\}}^*$ over $A \cup \{X\}$ where $A \in \mathcal{C}$ and $X \notin A$. $\pi_{A \cup \{X\}}^*$ can be computed by *collecting* the variable X from a clique B where $X \in B$. This operation, in principle, proceeds by recomputing the messages passed from B to A without eliminating X in the process. By exploiting the decomposition of clique and separator potentials induced by LARP, it is necessary only to consider a single potential of each message passed between cliques A and B. Only a single potential in each message contains the elimination of X in its evaluation.

Let $W \subseteq A \cup B$ be a subset such that there exists no $C \in \mathcal{C}$ with $W \subseteq C$. The potential $\pi_W$ can be computed by collecting variables to any clique in a subtree $\mathcal{T}_W$ of $\mathcal{T}$ spanning a set of cliques $\mathcal{C}_W$ such that $W \subseteq \bigcup_{C \in \mathcal{C}_W} C$.

**Theorem 6.1 [Variable Propagation]**
*Variable propagation of X from clique $B \in \mathcal{C}$ to clique $A \in \mathcal{C}$ where $X \notin A$ and $X \in B$ results in $\pi_{A \cup \{X\}}^*$ as the clique potential of A.*

**Corollary 6.2**
*Variable propagation of X from clique $B \in \mathcal{C}$ to clique $A \in \mathcal{C}$ where $X \notin A$ and $X \in B$ results in $\pi_{D \cup \{X\}}^*$ as the clique potential for each clique D on the path between B and A.*

Variable propagation is useful when determining multi dimensional partial derivatives as the following example illustrates.

**Example 6.3**
*Let A and B be adjacent cliques as in Figure 1 with potentials $\pi_A = (\Phi_A)$ and $\pi_B = (\Phi_B)$, respectively. Assume $\mathcal{T}$ is consistent, $\Theta_{X_i \mid \pi(X_i)} \in \Phi_B$, $\Theta_{X_j \mid \pi(X_j)} \in \Phi_A$, $\mathrm{fa}(X_i) \not\subseteq A$, and $\mathrm{fa}(X_j) \not\subseteq B$. Assume we need to determine the partial derivative $\frac{\partial^2 P(\epsilon)}{\partial \theta_{x_i \mid \pi_i} \partial \theta_{x_j \mid \pi_j}}$.*

*Variable propagation of $\mathrm{fa}(X_i)$ amounts to recomputing all messages of $\pi_{B \to A}$ created by elimination of some variable in $\mathrm{fa}(X_i)$. After variable propagation of $\mathrm{fa}(X_i)$ to A, we may compute:*

$$\frac{\partial^2 P(\epsilon)}{\partial \theta_{x_i \mid \pi_i} \partial \theta_{x_j \mid \pi_j}} = \left[ \sum_{A \setminus (\mathrm{fa}(X_i) \cup \mathrm{fa}(X_j))} c(\pi_A^* \ominus \pi_{X_i} \ominus \pi_{X_j}) \right] (x_i, \pi_i, x_j, \pi_j),$$

*where $\pi_A^*$ is the potential at A after variable propagation. In essence, we have propagated the CPD $\Theta_{X_i \mid \pi(X_i)}$ to A (and possibly other CPDs and potentials as well).*

## 7 A CAUTIOUS PROPAGATION SCHEME

In this section we present a cautious propagation scheme (Jensen 1995) of LARP referred to as cLARP. An early, simpler variant of cautious LP using Variable Elimination for message computation was introduced by Madsen and Jensen (1999).

The aim of cautious propagation is to support efficient calculation of the probability of subsets of the evidence by adjusting the message passing scheme. cLARP turns out to be quite useful for computing additional single and multi dimensional partial derivatives w.r.t. to evidence $\epsilon$ and subsets of $\epsilon$.

cLARP is based on *cautious entering of evidence* (Jensen 1995; Madsen and Jensen 1999): postpone the instantiation of an evidence variable X until the point where X would be eliminated by summation, if it had not been observed. This is contrary to the normal scheme where evidence functions are applied immediately to reduce the domain sizes of potentials.

Cautious entering of evidence implies that clique potentials are never changed (not even to reflect $\epsilon$) as evidence functions are only applied when calculating separator messages. Consider the calculation of $\pi_{A \to B}$ in (2). An evidence function $f(X)$ reflecting hard evidence on any variable $X \in A$, $X \notin B$ is applied to potentials of $\pi = \pi_A \otimes \pi_{\epsilon_A} \otimes \left( \otimes_{C \in \mathrm{adj}(A) \setminus \{B\}} \pi_{C \to A} \right)$ prior to variable elimination. In cLARP the projection operation of (2) proceeds as follows:

1. Let $\Phi$ be the potentials of $\pi$ where:
$$\pi = \pi_A \otimes \pi_{\epsilon_A} \otimes \left( \otimes_{C \in \text{adj}(A) \setminus \{B\}} \pi_{C \to A} \right).$$

2. Remove potentials of barren variables from $\Phi$.

3. If $X \in A \setminus B$ is observed, then apply $f(X)$ on potentials of $\Phi$.

4. Eliminate each non-barren variable $X \notin S$ from $\Phi$ by a sequence of arc-reversals and barren variable eliminations. Let $\Phi^*$ be the result.

5. Associate the message $\pi_{A \to B} = (\Phi^*)$ with S.

Cautious entering of evidence allows us to easily retract evidence, which again allows us to compute additional partial derivatives after a single propagation. For each clique $A$ and each separator $S$, we may compute single and multi dimensional partial derivatives for subsets $\epsilon' \subseteq \epsilon$ meeting at $A$ and $S$, respectively.

**Example 7.1**
*Figure 3 shows content of clique potential $\pi_A$ and messages $\pi_{C_0 \to A}$, $\pi_{C_1 \to A}$, and $\pi_{C_2 \to A}$ passed to A from adjacent cliques $C_0$, $C_1$, and $C_2$ (we do not include the evidence functions in the clique potentials shown (except for $f(X_3)$)). We assume evidence $\epsilon = \{x_1, x_3, x_7, x_8, x_9, x_{10}, x_{11}\}$.*

*At clique A we can compute partial derivatives of subsets of $\epsilon$ where any combination of $\{x_9, x_{10}, x_{11}\}$ and $\{x_3\}$ is retracted. For instance, we may compute:*

$$\frac{\partial P(\epsilon')}{\partial \theta_{x_4|x_2,x_3}} = \left[ c(\pi_A^* \ominus \pi_{X_4} \ominus \pi_{f(X_3)}) \right](x_2, x_3, x_4),$$

*where $\epsilon' = \epsilon \setminus \{x_3\}$.*

In cLARP we maintain two sets of separator mail boxes in the junction tree $\mathcal{T}$. One set of mail boxes for messages with evidence and one set of mail boxes for messages without evidence. We let $\mathcal{T} = (\mathcal{C}, \mathcal{S})$ denote the junction tree spanned by $\mathcal{C}$ and the separators $\mathcal{S}$ with evidence and let $\mathcal{T}' = (\mathcal{C}, \mathcal{S}')$ denote the junction tree spanned by $\mathcal{C}$ and the separators $\mathcal{S}'$ without evidence. Initially, we establish equilibrium of $\mathcal{T}'$ and next the equilibrium of $\mathcal{T}$ is established.

**Example 7.2**
*Since $\mathcal{T}'$ gives access to a set of virgin (clique) and separator potentials we may compute partial derivatives for additional subsets of the evidence. For instance:*

$$\begin{aligned}\frac{\partial^2 P(\epsilon')}{\partial \theta_{x_3} \partial \theta_{x_4|x_2,x_3}} &= \left[ c(\pi_A^* \ominus \pi_{C_0 \to A} \otimes \pi'_{C_0 \to A} \ominus \pi_{f(X_3)} \right.\\ &\qquad \left. \ominus \pi_{X_3} \ominus \pi_{X_4}) \right](x_2, x_3, x_4) \\ &= \left[ \phi'_2 \phi_7 \phi_8 \phi_9 \phi_{10} \phi_{11} \right](x_2, x_3, x_4),\end{aligned}$$

*where $\epsilon' = \epsilon \setminus \{x_1, x_3\}$ and $\pi'_{C_0 \to A} = (\{\phi'_2(X_2)\})$ is the message from $C_0$ to A in $\mathcal{T}'$.*

*In fact we may at A compute partial derivatives w.r.t. a large number of subsets of the evidence. The following sets can be retracted individually and in any combination $\{x_9, x_{10}, x_{11}\}$, $\{x_1\}$, $\{x_8\}$, $\{x_3\}$, and $\{x_7\}$.*

*Notice that even though $\{x_9, x_{10}, x_{11}\}$ and $\{x_7\}$ are contained in the same separator potential $\pi_{C_2 \to A}$, they can be retracted independently as the two subsets of evidence are independent (and because of the decomposition of $\pi_{C_2 \to A}$). This is neither possible in Shenoy-Shafer nor HUGIN propagation.*

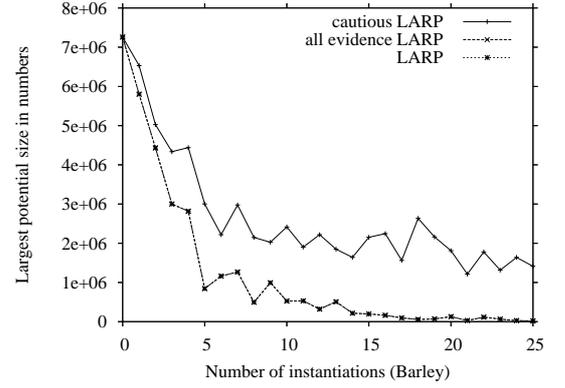

Figure 4: Average largest potential size in numbers.

## 8 PERFORMANCE ANALYSIS

The goal of our performance analysis is to empirically determine the overhead in computations introduced by propagating the entire set of evidence as well as the cautious propagation scheme. We analyzed both randomly generated and real-world BNs. For each network we propagated 25 randomly generated sets of evidence of size $n$ for $n = 0, \ldots, 25$.

Table 1: Information on the networks considered.

| Network | $|\mathcal{X}|$ | $|\mathcal{C}|$ | $\max_{A \in \mathcal{C}} s(A)$ | $\sum_{A \in \mathcal{C}} s(A)$ |
|---|---|---|---|---|
| Barley | 48 | 36 | 7,257,600 | 17,140,796 |
| ship-ship | 50 | 35 | 4,032,000 | 24,258,572 |

Due to space limitations we only report on the analysis performed on the *Barley*[1] and the *ship-ship* (Hansen and Pedersen 1998) networks. The results for these networks are representative for the networks we have analyzed. Some statistics for the networks are shown in Table 1. In the table, $s(A) = \prod_{X \in A} \|X\|$ is the state space size of clique $A \in \mathcal{C}$ where $\|X\|$ denotes the state space size of $X$ and $\mathcal{C}$ is the set of cliques of the junction tree.

---
[1] The Barley network can be downloaded from the homepage of the Department of Computer Science at Aalborg University: http://www.cs.aau.dk

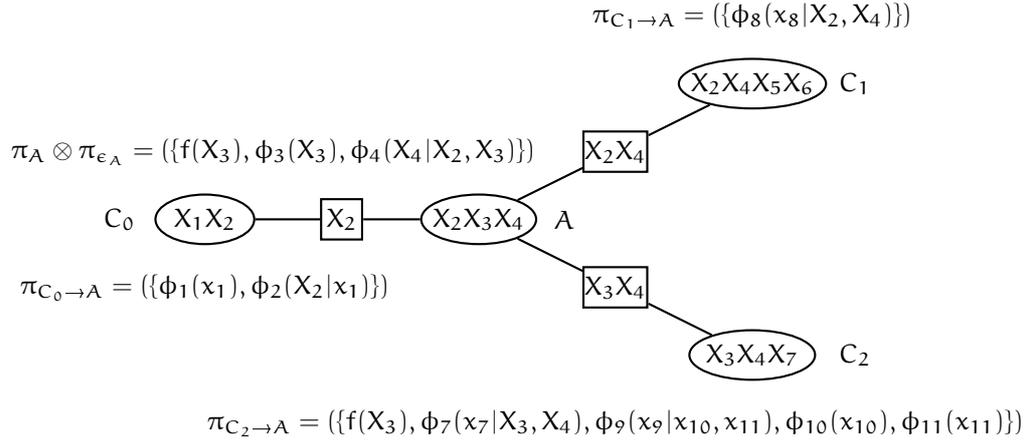

Figure 3: Content of the messages passed to clique A in Figure 2.

evidence), we introduce variable propagation as part of LARP, and finally we define the cLARP scheme.

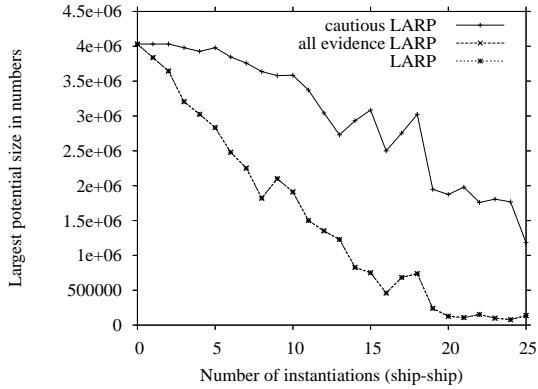

Figure 5: Average largest potential size in numbers.

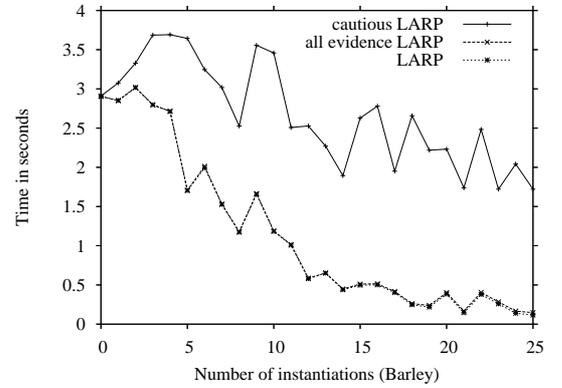

Figure 6: Average time in seconds for inference.

Figures 4 and 5 show the average size of the largest potential created during inference, while Figures 6 and 7 show the average time for inference in seconds. The state space size $s(\phi)$ of a potential $\phi$ is defined as $s(\phi) = \prod_{X \in \text{dom}(\phi)} \|X\|$.

cLARP consistently produce an average largest potential size that is at least as high as the average largest potential size of the all-evidence and normal propagation schemes. The cost of propagating all evidence is in most cases negligible.

## 9 DISCUSSION

This paper describes how single and multi dimensional partial derivatives may easily be computed from a LARP equilibrium. To facilitate the calculation of partial derivatives, we introduce a few modifications and extensions to LARP. We adjust the absorption algorithm to include all evidence (not only the relevant

The goal of the performance analysis was to determine the impact of propagating all evidence and the cLARP scheme. We are the first to demonstrate the impact of the cautious propagation scheme on the performance of inference. The experiments indicate that cLARP has a higher computational cost than LARP. This is expected as the cLARP scheme postpones the instantiation of an observed variable $X$ until the point where $X$ would be eliminated if it had not been observed.

That is, each observed variable $X$ is instantiated in each clique $A$ where $X \notin C$ and not instantiated in each clique $B$ where $X \in B$. Hence, when eliminating an unobserved variable $Y$ from a potential $\phi$ where $X \in \text{dom}(\phi)$ we are summing over $\text{dom}(\phi)$ instead of $\text{dom}(\phi) \setminus \{X\}$.

The experiments illustrates the computational overhead introduced by propagating all evidence and the cLARP scheme. The computational overhead of propagating all evidence is low, whereas the computational overhead of the cautious propagation scheme is higher. Notice that in some cases *all evidence* LARP and LARP have the same performance. This implies, for instance, that the average size of the largest potential created during inference is not impacted by propagating all evidence. In addition, since propagating all evidence does not change the time performance significantly either it suggests that all or almost all evidence is already propagated by LARP.

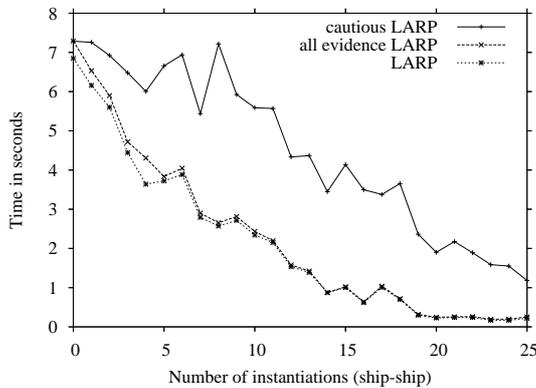

Figure 7: Average time in seconds for inference.

The LARP and cLARP algorithms give support for computing additional single and multi dimensional partial derivatives over the Shenoy-Shafer and HUGIN algorithms. In particular, we may compute partial derivatives w.r.t. subsets of separator messages and subsets of the evidence.

The decomposition of clique and separator potentials combined with the property that each probability potential has (at most) one head variable suggest that the differential semantics of LARP is well-suited for hypothesis driven sensitivity analysis. Future work includes an application of the differential semantics of LARP to hypothesis driven parameter sensitivity analysis. Similarly, cLARP supports easy calculation of a large set of single and multi partial derivatives of the evidence and subsets of the evidence. For this reason, the architecture should be well-suited for hypothesis driven parameter sensitivity analysis.